# Why Blocking Targeted Adversarial Perturbations Impairs the Ability to Learn

Ziv Katzir[1], Yuval Elovici[2]

Department of Software and Information Systems Engineering,
Ben-Gurion University of the Negev, Beer-Sheva, Israel
[1]zivka@post.bgu.ac.il, [2]elovici@bgu.ac.il

**Abstract.** Despite their accuracy, neural network-based classifiers are still prone to manipulation through adversarial perturbations. Those perturbations are designed to be misclassified by the neural network, while being perceptually identical to some valid input. The vast majority of attack methods rely on white-box conditions, where the attacker has full knowledge of the attacked network's parameters. This allows the attacker to calculate the network's loss gradient with respect to some valid input and use this gradient in order to create an adversarial example. The task of blocking white-box attacks has proven difficult to solve. While a large number of defense methods have been suggested, they have had limited success. In this work we examine this difficulty and try to understand it. We systematically explore the abilities and limitations of defensive distillation, one of the most promising defense mechanisms against adversarial perturbations suggested so far in order to understand the defense challenge. We show that contrary to commonly held belief, the ability to bypass defensive distillation is not dependent on an attack's level of sophistication. In fact, simple approaches, such as the Targeted Gradient Sign Method, are capable of effectively bypassing defensive distillation. We prove that defensive distillation is highly effective against non-targeted attacks but is unsuitable for targeted attacks. This discovery leads us to realize that targeted attacks leverage the same input gradient that allows a network to be trained. This implies that blocking them will require losing the network's ability to learn, presenting an impossible tradeoff to the research community.

## 1.  Introduction
The discovery of adversarial perturbations against neural networks [1] unleashed an arms race between attack and defense methods in recent years. In this race, ever more powerful attack methods were discovered, creating adversarial examples that are perceptually indistinguishable from valid input, yet are incorrectly classified by the neural network. Most of the attack methods suggested so far operate in white-box conditions and assume that the adversary has full knowledge of the attacked network's parameters. This knowledge allows the attacker to calculate the network's loss gradient with respect to some valid input point and use this gradient in order to perturb the input until it is able to mislead the classifier. This process is quite similar to the "standard" gradient back-propagation performed during network training. However, in this case the network's weights are kept constant, and the input itself is being updated. The result is an adversarial example that is misclassified by the neural network.

Throughout most of this arms race attackers have had the upper hand. The task of blocking adversarial perturbations, or even distinguishing them from valid input, has proven difficult. A large number of defense methods have been suggested, but they have had very limited success [16][17]. The sheer number of failed attempts suggests that there is some fundamental reason that makes adversarial perturbations so difficult to defend against. In spite of this situation, only a limited amount of research has been performed in an attempt to understand the root cause of this difficulty. In this work we aim to shed light on the fundamental principles that make defense mechanisms against adversarial perturbations fail time after time. We focus our work on white-box attack methods, attempting to understand the nature of the loss gradients that lead to their creation.

We start our analysis by examining defensive distillation [6], one of the more promising defense methods suggested so far. Defensive distillation is particularly interesting in our case as it attempts to nullify the loss gradient resulting from the last softmax layer and thus prevents the creation of adversarial examples. In order to do so, defensive distillation increases the probability estimate associated with the most probable class and decreases the probabilities of all other classes. This leads to a reduction of several orders of magnitude in the loss gradient associated with the most probable class, until, in many cases, it becomes too small to be represented by a standard 32-bit floating point variable. In doing so, defensive distillation leverages the floating point accuracy limitation in order to block white-box attack methods.

The concept of gradient masking [19], as implemented by defensive distillation, appeared to be highly promising. However, not long after defensive distillation was published, Carlini and Wagner (C&W) [5] designed a highly sophisticated attack method in order to overcome it. The C&W attack ignores the classifier's softmax layer, and introduces a dedicated set of a loss function and an optimization scheme. Together, those elements are able to "restore" the lost gradient signal and use it for creating adversarial examples.

The C&W attack is commonly credited with being the first to bypass defensive distillation, and its success is attributed to its high level of sophistication. However, our work indicates that this perception is in fact inaccurate. We follow the footsteps of [8] and show that defensive distillation can be bypassed using simple attack methods, such as the Targeted Gradient Sign Method (TGSM) [7]. In order to

4explain this observation, we present a simple framework for visualizing the outputs of the last two layers of the classifier network. We do that by reducing a multi-class classification task into a binary problem, arbitrarily choosing two of the original classes. The reduced problem domain allows us to visualize the output of the last two layers of the classifier network using a two dimensional chart. Based on our visualization framework, we can explore the influence of defensive distillation over the neural network's gradients when faced with different types of adversarial attacks. This exploration led us to hypothesize and eventually prove that defensive distillation is highly effective against non-targeted attacks but is unsuitable for handling targeted attacks. This alone is a valuable insight, however it leads us to a considerably more important understanding, which is the main contribution of this work: White-box targeted adversarial attacks leverage the same input gradients that are used for training a classifier network. Therefore, the ability to block them relies on making the classifier "untrainable", presenting an impossible tradeoff to the research community.

The remainder of this paper is organized as follows: Section 2 provides background about adversarial manipulations and defense mechanisms. Section 3 contains a description of our gradient analysis framework and the use of this framework to analyze the function of defensive distillation in the context of various attack methods. Section 4 provides a formal proof that defensive distillation is effective against non-targeted attack methods but unsuitable for handling targeted ones. Section 5 provides the main contribution of this work. It explains why blocking targeted adversarial perturbations requires eliminating the network's ability to learn. Finally, Section 6 provides our final conclusions and future research avenues.

## 2. Background

In this section, we provide the needed background about adversarial manipulations and survey key attack and defense methods.

### 2.1. Crafting Adversarial Examples

Crafting adversarial examples in white-box conditions generally implies solving a set of two constraints. The attacker aims to identify some small perturbation $\delta$, that when added to a valid input vector $x$, will cause the classifier $f(\cdot)$ to misclassify the perturbed input. This perturbation must also be small enough to ensure it is undetectable by humans. Using $x'$ to denote the resulting adversarial example $x' = (x + \delta)$, and $Y$ to denote the true class label, we can formally define the adversary's goal as follows:

$$f(x') \neq Y \tag{1}$$

$$s.t. \|\delta\|_p < \varepsilon$$

The first constraint in equation (1) ensures incorrect classification, while the second controls the perturbation magnitude. Here $\varepsilon$ stands for the perturbation radius given some context specific distance metric $\|\cdot\|_p$. Limiting the value $\varepsilon$ aims to prevent the perturbation from being detected.

Various optimization methods for solving the constraints in (1) have been suggested in recent years. Typically, the methods include computation of $f(x)$, calculation of the loss gradient with respect to the input, and one or more steps in which the input is perturbed based on the gradient, in order to cause it to be misclassified by the network.

Attack methods are often divided into two classes: targeted and non-targeted attacks. In the case of non-targeted attacks, the attacker wishes to deviate from the true class label but does not care which other class is chosen. This goal is typically achieved by maximizing the network's loss with respect to the true class.

Targeted attacks, on the other hand, aim to manipulate the network to predict some specific class label given the input. Like non-targeted attacks, those methods use the loss gradient in order to perturb the input. However, instead of maximizing the loss with respect to the true class, targeted attack methods aim to minimize the loss associated with the selected target class. By doing so, targeted attack methods allow the attacker full control of the classification result.

Much of the prior research of adversarial examples deals with image classification tasks. In this context, three distance metrics are commonly used as a proxy for human perception: 1) $L_0$ measures the number of perturbed features (i.e., image pixels), 2) $L_2$ measures the perturbation's Euclidian norm, and 3) $L_\infty$ measures the maximal change to any of the input features. Applying adversarial manipulations to additional content domains similarly requires the identification of a suitable distance metric.

The following sub-sections provide details about several notable attack methods and describe the effect of adversarial perturbations on the classifier network.

**2.1.1. Fast Gradient Sign Method and Basic Iterative Method**

The Fast Gradient Sign Method (FGSM) [3] was the first identified method to allow computationally efficient crafting of non-targeted adversarial examples. Using a single gradient calculation, the algorithm perturbs each of the input features by a magnitude of $\varepsilon$ in the direction of the loss gradient. Formally this is expressed as:

$$x' = x + \varepsilon * sign(\nabla_x J(x, Y)) \qquad (2)$$

where $J(x, Y)$ represents the classifier's loss given an input vector $x$ and true class label $Y$.

The Basic Iterative Method (BIM) [4] is an extension of FGSM. Instead of performing a single step in the direction of the gradient, BIM performs up to $h$ iterations of equation (2). Each iteration is using a smaller perturbation step $\gamma$, such that $\varepsilon = h * \gamma$. The gradient is recalculated in each step, and the algorithm stops as soon as the perturbation is able to mislead the classifier. The result, in most cases, is a considerably refined perturbation pattern compared to FGSM.

**2.1.2. Targeted Gradient Sign Method**
FGSM was originally designed as a non-targeted attack method. It causes input to be misclassified by the classifier network but does not allow the attacker control of the classification output. In [7], the authors presented a targeted variant of FGSM referred to as the Targeted Gradient Sign Method (TGSM), as well as a targeted variant of BIM. Instead of maximizing the loss with respect to the true class label, TGSM attempts to minimize the loss with respect to the target class $l_{target}$ (note how the perturbation is subtracted from the valid input) :

$$x' = x - \varepsilon * sign\left(\nabla_x J(x, l_{target})\right) \qquad (3)$$

**2.1.3. Jacobian Saliency Map Attack (JSMA)**
In [2], the authors suggested a greedy iterative algorithm for crafting targeted adversarial examples. In each iteration, the algorithm computes a saliency map based on the network's Jacobian and perturbs the two most salient input features. One of the features is used for decreasing the probability estimate of the correct class, while the other is used for increasing the probability estimate of the target class. This procedure minimizes the number of perturbed features.
Indeed, the iterative nature of the JSMA algorithm minimizes the number of perturbed features, however computing the full Jacobian is highly demanding in terms of computation resources. These high computation demands make JSMA impractical for networks with a high input dimensionality, such as ImageNet.

**2.1.4. Carlini & Wagner (C&W) Attack**
In [5], the authors presented an attack method that can to this day defeat most, if not all, known defense mechanisms. The C&W attack was specifically designed to bypass defensive distillation [6], which was considered an unbeatable defense mechanism at the time.
Defensive distillation aims to prevent the adversary from using the network's loss gradient for creating adversarial examples. It does so by eliminating the loss gradient resulting from the network's softmax layer. In order to overcome the distillation effect, the C&W attack introduces two main concepts: First, it ignores the softmax output and instead uses the output of the linear layer that immediately precedes it, known as the logits layer; then, it replaces the network's original loss function with an attack oriented one and introduces a dedicated optimization process. This process allows a tradeoff between attack success and the size of the resulting perturbation.
The same two principles are applied with minor modification to form a set of three attack methods (one for each commonly used distance metric), all sharing the same iterative optimization scheme.
The authors started by rephrasing the objective (loss) function used for crafting a targeted attack. Instead of requiring that

$$f(x + \delta) = l_{target} \qquad (4)$$

where $l_{target}$ is the label of a destination target class, they introduced a new objective function $q(x + \delta)$ so that

$$f(x + \delta) = l_{target} \leftrightarrow q(x + \delta) \leq 0 \qquad (5)$$

The original objective function presented in equation (4) is highly nonlinear, making it difficult to solve. However, using a monotonically increasing function for $q$, the authors were able to construct an equivalent optimization problem that is both easier to solve and controls the tradeoff between perturbation size and the need to mislead the classifier:

$$\min \left( \|\delta\|_p + c \cdot q(x + \delta) \right) \qquad (6)$$
$$s.t. \quad x + \delta \in [0,1]^n$$

The authors used a standard gradient descend optimizer in order to solve the minimization problem presented in equation (6). The constant $c$ is used for balancing the effect of the perturbation norm against the value of the function $q$. Its goal is to ensure that both terms influence the optimization process equally. The result is a powerful set of attacks for which there are no known defense or detection mechanisms.
The C&W attack is commonly credited for being the first attack method to overcome defensive distillation. In this work, we show that much simpler attack methods, such as TGSM, are in fact capable of bypassing defensive distillation as well. Moreover, we prove that this ability is not related to the level of attack sophistication, but rather to the targeted nature of those attacks.

### 2.2. Defense Mechanisms

#### 2.2.1. Adversarial Training
Adversarial training [1][3] is perhaps the most immediate line of defense against adversarial manipulations. It is based on the intuition that adversarial examples occur in sections of the input space that are underrepresented in the training data. Adversarial training therefore wishes to augment and enrich the training set. It is based on iteratively training the classifier network using adversarial examples by 1) training a network to be sufficiently accurate with normal input, 2) generating adversarial examples, 3) augmenting the training input, and 4) fine-tuning the classifier.

This simple approach has demonstrated greater model resilience than undefended classifiers, however it has a few shortcomings: 1) it is difficult to scale to classifiers that process high resolution input [4] like the ImageNet dataset, 2) adversarial training based on weak attacks does not provide an adequate defense against stronger attacks [11], and 3) it is fairly easy to construct adversarial examples against a network that has already been trained to cope with some adversarial examples [12].

**2.2.2. Defensive Distillation**

The term distillation [18] refers to the process of training one network using the softmax outputs of another network. Originally, this process was aimed at reducing the computational resources required for using a neural network. Hence, the distilled network included considerably fewer neurons. Distillation works by first training a "teacher" model using the ground truth labels; then using the trained teacher model to predict the probability of each training example to belong to each of the potential class labels; and finally, using this set of probability estimates to train a "student" model. The main goal of this process is to provide the student model with knowledge about the interaction between different class labels, which is much richer information than a simple indication of the correct class.

Defensive distillation [6] adapts the basic distillation process described above in order to increase the resilience of the distilled network against adversarial examples. Defensive distillation does not attempt to reduce the computational load of using a neural network, hence the teacher and student network models share a common architecture. Instead, it attempts to eliminate the loss gradient propagated from the student network's softmax layer. In order to do so, defensive distillation increases the size of the logit component associated with the most probable class compared to the components associated with all other classes. This in turn decreases the size of the loss gradient of the true class by several orders of magnitude until in many cases it can no longer be represented by a standard 32-bit floating point variable. This phenomenon, which is referred to as gradient masking [19], prevents the adversary from using the loss gradient to create adversarial examples.

In order to create the gradient masking effect, defensive distillation applies a minor modification to the softmax calculation of both teacher and student networks. The standard softmax equation is modified to account for the distillation temperature $T$ as follows:

$$Softmax(z(x),T)_i = \frac{e^{z_i/T}}{\sum_{j=0}^{N-1} e^{z_j/T}} \quad (7)$$

Here, $N$ denotes the number of potential classes identified by the network, $i \in [0, N-1]$ is a specific class label index, and $z(x) = (z_0, z_1, ..., z_{N-1})$ represents the output of the network's last linear layer known as the logits layer.

As the distillation temperature increases, the value of $T$ becomes larger than $z_i$, causing each of the softmax outputs approach $1/N$. In such a case the softmax assigns equal probability to all potential classes and is therefore unable to predict the true class label. In order to allow correct classification, the training process of the distilled network is therefore required to compensate for the training temperature by increasing the magnitude of the logits layer's output. Training a network with a high distillation temperature therefore, causes an increase in the magnitude of the logits that is proportional to the distillation temperature.

In the case of defensive distillation both the teacher and student networks are trained using a high distillation temperature (e.g., 30). However, when the student model is used for inference, the temperature is set to one. The logit signal is still increased proportionally to the distillation temperature used during training, however setting the temperature to one reverts equation (7) back to a standard softmax. Therefore, the softmax calculation resembles a hard max. Informally, we say that the classifier becomes more certain about its classification output, making it more resilient to adversarial perturbations.

At the time of its discovery, defensive distillation was able to defeat all known adversarial attack methods. The C&W attack [5] was designed to bypass defensive distillation, and it was considered the first successful attempt to do so. However, as we show in this work, simpler and earlier attack methods are also able to bypass defensive distillation. In fact, defensive distillation is inherently incapable of defending against targeted attack methods, including (but not limited to) the C&W attack.

### 3. Exploring Classifier Network Input Gradients

In [8], the authors show that TGSM can bypass defensive distillation. Through a series of experiments, they provide empirical evidence that defensively distilled networks eliminate the input gradients associated with FGSM perturbations, but are unable of doing the same in the case of TGSM. Despite observing this difference, the authors were unable to explain it.

We follow up on this work, providing a formal analysis of the classifier's input gradients and ultimately proving that defensive distillation is highly effective in defending against non-targeted attacks, but is unsuitable for handling targeted attacks.

We start by reducing the MNIST digit classification task [9] into a binary classification problem, and training a suitable classifier network. First, two out of the ten available classes are chosen at random. Then the original dataset is filtered to include only relevant samples for both training and testing data. Finally a simple convolutional neural network is trained, based on the network architecture listed in [15]. Following common practices in neural network based classification, the network uses the ReLU activation function, a final softmax output and the cross-entropy loss metric.

Reducing the number of classes into two allows us to visually explore different aspects of the classifier network's operation using two-dimensional charts. As a first experiment, we leverage this ability for studying the logit values produced by the network for both normal and adversarial input. Starting with non-targeted FGSM and plotting normal logit values side by side to the adversarial ones, the effect of adversarial examples over the classifier network is highly evident.

Figure 1 illustrates the result of this experiment, comparing the logit values for normal and FGSM perturbed inputs. The two logits components are indicated by the two axes of the plot. The horizontal axis represents the logit component associated with class 0, while the vertical axis represents the logit component of class 1. Each plotted point represents the logit values of a single input image from the test set, all originally belonging to class 1. Given that softmax preserves does not change the size order of its outputs compared to its inputs, we get that in order to be correctly classified, the vertical component of a given data point should be larger than its horizontal component. This is indeed the case for most unperturbed inputs, indicating that our two-class problem is rather straightforward.

The effect of FGSM perturbation is also clearly evident in Figure 1. For the vast majority of the perturbed inputs the horizontal component becomes larger than the vertical one causing the data point to be incorrectly classified. Not surprisingly, a similar looking chart is produced when using other attacks such as JSMA, BIM, TGSM, targeted BIM, and C&W.

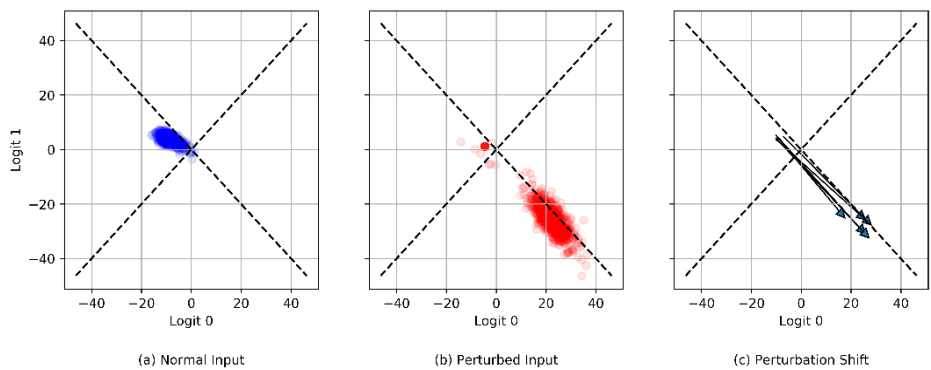

**Fig. 1.** The effect of FGSM on the logit layer outputs. The horizontal and vertical axes stand for logit components 0 and 1 respectively. (a) Logit values given normal input, (b) logit values for FGSM perturbed input, (c) perturbation shift, comparing the original and perturbed input.

We continued, performing a second experiment by applying FGSM perturbations to a defensively distilled version of the classifier network. Figure 2 presents the results of this experiment using a distillation temperature of $T = 30$. The following key observations can be made by analyzing Figure 2:

- Defensive distillation is highly effective against FGSM. This is evident by comparing Figure 2 (a), and (b). For most of the plotted data points the logit component associated with class 1 remains bigger than that of class 0, indicating they are correctly classified despite the attack.
- The absolute values of the logit components for the defensively distilled network are roughly 15 times larger than the non-distilled network. This is a by-product of training a neural network with defensive distillation. The network is "forced" to compensate for the distillation temperature by strengthening the logit signal.
- Non perturbed data points for the defensively distilled network align closely to the $X = -Y$ line. Moreover, data points are generally pushed farther away from the origin in the defensively distilled network compared to the non-distilled network. This is an illustration of the fact that distilled networks are "more certain" about their classification outputs.

By applying the JSMA and BIM attack methods to the distilled network we have observed similar outcomes. This indicates the effectiveness of defensive distillation against those attack methods, and confirms previously reported results.

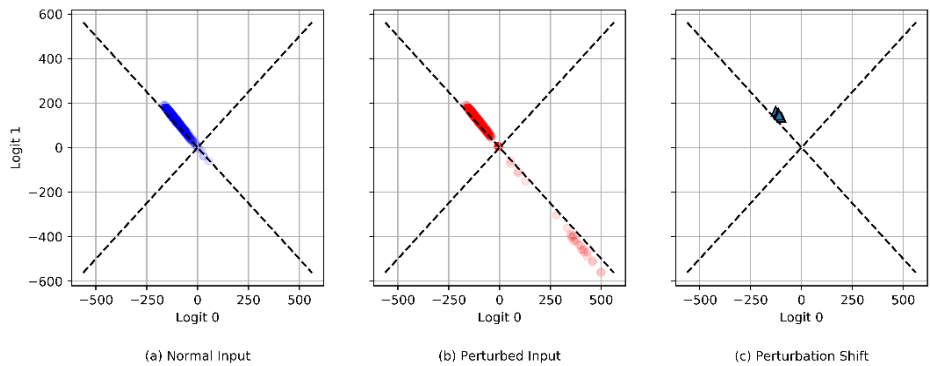

**Fig. 2.** The effect of FGSM on a defensively distilled classifier network. (a) Logit values given normal input, (b) logit values for FGSM perturbed input, (c) perturbation shift, comparing the original and perturbed input.

We continued to explore the defensively distilled classifier network using TGSM, targeted BIM, and the C&W attacks. In all three cases defensive distillation was unable to block the perturbation. Logit values were manipulated and an incorrect classification result was chosen. Figure 3 presents the results of this experiment using the TGSM attack. While this outcome was expected in the case of the C&W attack, it was unexpected in the case of TGSM and targeted BIM attacks. Recall that the C&W attack is considered the first method to successfully bypass defensive distillation, and that its success is attributed to its sophistication. Both TGSM and targeted BIM were published prior to the C&W attack and are far simpler to implement. In addition, we did not expect to see a difference between the targeted and non-targeted attack variants in our case. In the context of our binary classification problem, we assumed that targeted and non-targeted attacks are practically equivalent. Preventing correct classification for one class should implicitly mean that the input is classified to the other class. Therefore, the different behavior observed for targeted and non-targeted attacks requires further investigation.

After observing that simple, targeted attack methods (e.g., TGSM and targeted BIM) are able to bypass defensive distillation, we recalled that the C&W attack is targeted by nature. In fact, this attack method does not have a non-targeted variant. This understanding led us to rethink the fundamental reasons that allow C&W to bypass defensive distillation in the first place. We hypothesized that defensive distillation is simply unsuitable for handling targeted attacks, while noting that JSMA, which is a targeted attack, is an exception that will require further investigation. We have further hypothesized that it is the targeted nature of those attacks and not their level of sophistication that allows them to bypass defensive distillation.

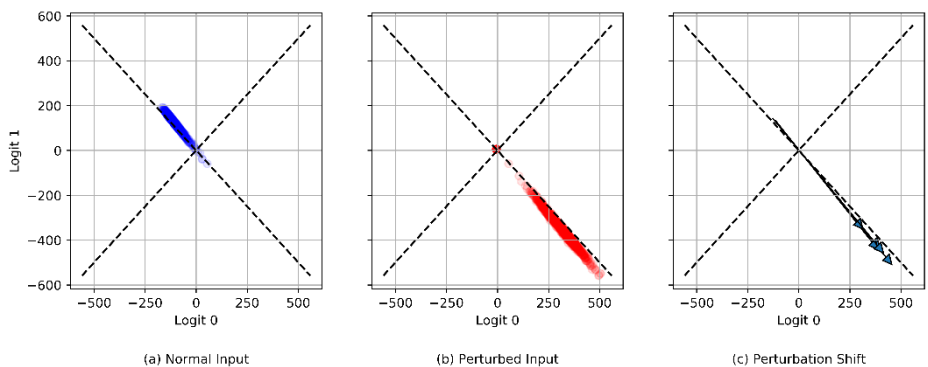

**Fig. 3.** The effect of TGSM on a defensively distilled classifier network. (a) Logit values given normal input, (b) logit values for TGSM perturbed input, (c) perturbation shift, comparing the original and perturbed input.

## 4. Formal Analysis of Input Loss Gradients

As indicated in Section 3, empirical exploration of the logit values of the defensively distilled two-class classifier led us to believe that defensive distillation is inherently unsuitable for dealing with targeted attacks. In this section, we provide a formal proof to support this hypothesis.

### 4.1. Non-Targeted Input Loss Gradients

Following our previously defined notation, we use $N$ as the number of different classes that can be assigned by our classifier network. We further use $P_i$ to denote the probability estimate assigned by the network to some class index $i \in \{0..N-1\}$ given an input sample $x$. Finally, we use $Y = (y_0, y_1, ..., y_{N-1})$ to represent the one-hot encoded ground truth classification of the input.

Under those notations, the network's cross-entropy loss for the input sample is expressed as:

$$L = -\sum_{i=0}^{N-1} y_i \log(P_i) = -\sum_{i=0}^{N-1} y_i \log(P_i) \\ = -\log(P_{true}) \quad (8)$$

By using $Z = (z_0, z_1, ..., z_{N-1})$ to denote the output of the logits layer, we can provide a formal definition of the loss gradient with respect to the i-th logit:

$$\frac{\partial L}{\partial z_i} = P_i - y_i \quad (9)$$

The complete proof of equation (9) is provided in Appendix A. Note that the gradient formulation presented in equation (9) is independent of the overall network architecture. It is the result of the gradients of the cross-entropy loss metric and softmax function only.

Using equation (9), we can see how defensive distillation decreases or even eliminates the input gradients in the case of non-targeted perturbations. Assuming a given input is correctly classified by the network, the increase in distillation temperature $T$ causes the network to increase the logit component associated with the true class label, $z_{true}$. As a result the probability estimate assigned by the network to the correct class label $P_{true}$ increases towards one.

Recalling that $Y$ is the one-hot representation of the ground truth label, we get that:

$$\lim_{T \to \infty} \frac{\partial L}{\partial z_{true}} = \lim_{T \to \infty} (P_{true} - y_{true}) = 1 - 1 = 0 \quad (10)$$

The loss gradient attributed to the logit value of the true class $z_{true}$ decreases towards zero in that case. When the training temperature is high enough, gradients will become too small to be represented by a standard 32-bit floating point variable.

Propagating the gradient backwards to the network's input will result in a decrease or even an elimination of the input gradient, making the classifier network resilient to non-targeted perturbations.

### 4.2. Targeted Input Loss Gradients

In the case of a targeted attack, loss is calculated with respect to the target class. Using $Y_{target} = (yt_0, yt_1, ..., yt_{N-1})$ to denote the one-hot encoded vector of the target class, we can update equation (9) to reflect the loss gradient of a targeted attack:

$$\frac{\partial L_{target}}{\partial z_i} = P_i - yt_i \tag{11}$$

In the case of a defensively distilled network, $P_{true}$ approaches one, causing $P_i$ to decrease towards zero for any other class. With that understanding, we can now express the limit of targeted loss gradient as distillation temperature is increased:

$$\lim_{T \to \infty} \frac{\partial L_{target}}{\partial z_{target}} = \lim_{T \to \infty} (P_{target} - yt_{target}) = 0 - 1 = -1 \tag{12}$$

The loss gradient with respect to the logit as expressed in equation (12) will hence approach zero for the true class label, or negative one (-1) for any other class.
Defensive distillation causes the absolute value of the input gradient with respect to the target class to be equal to one. This typically represents an increase of several orders of magnitude to the gradient, compared to the non-distilled case. This gradient makes defensive distillation inherently unsuitable for preventing targeted attacks.

### 4.3. The JSMA Exception

As noted earlier, JSMA is a targeted attack method. Given our analysis in Section 4.2, one might expect it to defeat defensive distillation. However, as reported in [5][6], this is not the case. Defensive distillation is highly effective in blocking JSMA-based perturbations.
Resolving this apparent contradiction requires some deeper understanding of the JSMA perturbation algorithm. JSMA is an iterative algorithm. In every iteration the algorithm attempts to perturb two input features based on the input-loss Jacobian. One feature is aimed to reduce the network's certainty of the true class, while the other is aimed to increase its certainty for the target class. From a gradient point of view, JSMA is essentially built out of a "non-targeted" part and a "targeted" part. The gradient used for reducing the network's certainty at the true class is what we refer to in Section 4.1 as a non-targeted gradient. As we have demonstrated, this gradient is nullified as a result of defensive distillation. JSMA is therefore unable to identify a suitable pair of input features to be perturbed, and thus terminates prematurely, without being able to form an adversarial example. JSMA's reliance on the classifier's softmax outputs does not allow it to overcome the gradient masking effect created by defensive distillation.

### 4.4. Analyzing Black-Box Attacks

Under black-box settings, the adversary is assumed to have no knowledge over the classifier network's parameters. The adversary is therefore unable to compute the network's loss gradient directly. Instead, he is required to use some proxy to either the gradient itself or the network as a whole. In [14] the authors discovered adversarial examples are transferrable between different classifier networks. They show it is possible to attack a classifier model using a surrogate one. An adversary can train a surrogate classifier network, use white-box attack algorithms against this surrogate and finally use the resulting adversarial examples in order to attack the original classifier. The authors further show that this process is highly effective even when the architecture and dataset used for training the surrogate model is different than the original one.

The authors of [14], further show that defensive distillation is ineffective in the context of black-box attacks. They train a surrogate non-distilled network, craft adversarial examples against this surrogate model using FGSM, and eventually feed the resulting adversarial examples to a distilled classifier network. As a result of this simple procedure, the defensively distilled network incorrectly classifies most of the adversarial examples presented to it. Based on those results, the authors assumed that defensive distillation is only effective in proximity to the data points used for training the model. By leveraging the results obtained in Section 4.2 we can provide a more accurate explanation – defensive distillation will eliminate the input gradient for all correctly classified inputs (whether used for training or not) but will not affect any incorrectly classified input.

Defensive distillation increases the probability estimate associated with the most likely class in favor of the others. For an input that is correctly classified by the non-distilled network, the most probable class is the true class ($P_{true}$). Increasing the distillation temperature will hence make this estimate approach one, causing the gradient expressed in equation (9) to decrease towards zero. However, when the original network fails to correctly classify an input, defensive distillation will only make the network more certain of the incorrect class label.

With that in mind, the role of the surrogate model becomes clear. The perturbed input points crafted using the surrogate model are incorrectly classified by the original classifier network and are hence highly likely to be incorrectly classified by the distilled network as well.

### 4.4.1. Optimizing Defensive Distillation

As described in Section 2.2.2, the defensive distillation process requires training of two distinct classifier networks. First a teacher model is trained, then a student model is trained based on the teacher's predictions. The two-phased training procedure was initially aimed to allow knowledge about the interactions between classes to be learned by the student model. However, as the distillation temperature increases, the probability estimate of the true class ($P_{true}$) approaches one, pushing all other probabilities closer to zero. In practice, increasing the distillation temperature causes the softmax function to be gradually transformed into a max function. The amount of such knowledge about cross-class interactions rapidly decreases making the two-phased training procedure rather meaningless. Figure 4 provides a graphical illustration of the effect of defensive distillation on the softmax output. Values for the defensively distilled network are set to either (1,0) or (0,1), demonstrating how softmax is transformed into a max function.

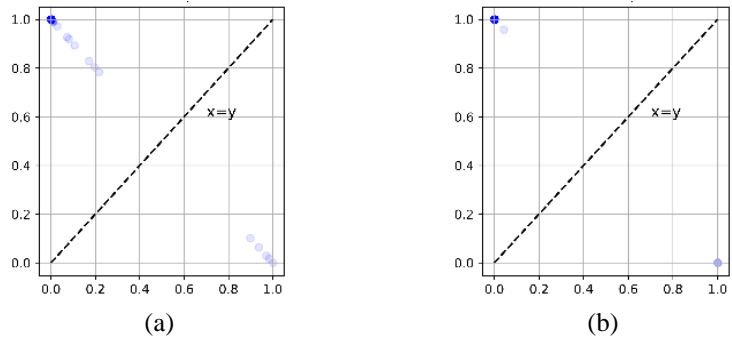

**Fig. 4.** Comparing (a) normal and (b) defensively distilled softmax values.

This analysis of the softmax outputs suggests it should be possible to optimize the defensive distillation process by using just a single network training phase. Training a single model with high distillation temperature in order to increase its certainty of the correct class label, and then setting the temperature to 1 during inference. In order to test the validity of this approach, we implemented an optimized version of the distillation process. Instead of the dual training procedure, we trained a single model using a high distillation temperature; set the distillation temperature to one; and used this model for inference. We have tested our approach with two defensively distilled networks trained using the MNIST and CIFAR-10 [13] datasets. Comparing classification accuracy as well as model resilience to adversarial examples, showed the optimized training procedure is on-par with the original defensive distillation.

Apart from the inability to cope with targeted attacks, the dual training process is considered the most problematic aspect of defensive distillation. Indeed, training can be considered a one-time investment of effort. However, training a modern neural network (e.g., ResNet) can take several days and require the combined computational power of many GPUs. Understanding that similar results can be achieved with a single training process hence represents a valuable optimization.

## 5. Blocking Targeted Adversarial Examples Implies Losing the Ability to Learn

The formal analysis provided in Section 4.2 leads to a much more significant insight about targeted white-box attacks against neural network classifiers. In order to fully appreciate the implications of this analysis, we should first recall a few of the fundamental facts regarding training a neural network classifier:

1) The order of classes within the network output is arbitrary. Choosing a different mapping of classes to class indexes will not affect the network's ability to train or the resulting accuracy.
2) Network loss is a monotonically increasing, non-negative function. It is equal to zero when the network predicts the correct class label, and increases in correlation with the classification error.
3) Classifier networks are trained using stochastic gradient descend starting from a random set of weights. In order to allow training to converge in a reasonable amount of time, a high loss gradient is required whenever classification error is high. Similarly low loss gradients are required when classification error is low, in order to be able to stop the training process.

Equipped with those three fundamental facts, we can now re-examine the gradient calculations provided in sections 4.1 and 4.2. Targeted attacks compute the loss with respect to some class other than the ground truth one. The loss function refers to this target class as if it was true, hence yielding a high loss value and a high loss gradient (points 2,3 above).

From a network's point of view, calculating the loss in the case of a targeted attack is similar to permuting the order of classes and attempting to retrain the network. In both cases the position of the "one" within the one-hot encoded labels vector is different than the position predicted by the network, leading to a high loss value. The nature of the loss function itself implies that in both cases a strong loss gradient is formed as well.

This understanding has profound implications on our ability to block targeted adversarial attacks. Realizing that those attacks leverage the same input gradient that allows a network to be trained, implies that blocking them will require losing the network's ability to learn. This paradox calls for a new approach to protection against targeted adversarial attacks.

## 6. Conclusions

In this paper, we aimed to fill the existing gap in understanding the fundamental reasons that make adversarial examples so difficult to defend against in white-box settings.

We started by examining defensive distillation, one of the most promising defense methods suggested so far. This method aims to eliminate the network's loss gradients, thereby preventing them from being used for creating adversarial examples. Using explicit formulation of the loss gradients resulting from the cross-entropy loss and softmax activation function, we proved that defensive distillation can provide complete protection against non-targeted attacks but is not suitable for targeted attacks.

While this understanding is, in itself valuable, it led to a much more significant insight about white-box adversarial attacks. The loss gradients used by targeted and non-targeted attack algorithms are in-fact different. Moreover, we showed that targeted adversarial attacks leverage the same gradients that are used for training the classifier network. This implies that in order to block targeted adversarial examples, the network's ability to learn must be disabled.

In addition to providing a solid theoretical explanation for the difficulty in constructing effective defense mechanism, this final conclusion points out a new path to pursue with regard to the study of defense mechanisms. It indicates that researchers should look for ways to train a network and then obstruct or impede its ability to learn without harming its classification accuracy.

# Appendix A – Derivation of the Cross Entropy Loss and Softmax Activation Function

This section describes the gradient derivation of the cross-entropy loss combined with the softmax activation function. This derivation was first formally described in [10]. Our version is based on [20] which provides an easier read.

Derivation of the Softmax Activation Function
Let $z_i$ denote the i-th component of the logits layer output given some network input $x$.
The probability estimate of the i-th class associated by the softmax function to that input can be expressed as:

$$P_i = \frac{e^{z_i}}{\sum_{k=0}^{N-1} e^{z_k}}$$

The derivative of $P_i$ with respect to $z_k$ can then be computed as follows:

$$\frac{\partial P_i}{\partial z_j} = \frac{\partial \left( \frac{e^{z_i}}{\sum_{k=0}^{N-1} e^{z_k}} \right)}{\partial z_j}$$

When $i = j$ we get:

$$\frac{\partial P_i}{\partial z_j} = \frac{\partial \left( \frac{e^{z_i}}{\sum_{k=0}^{N-1} e^{z_k}} \right)}{\partial z_j} = \frac{e^{z_i} \sum_{k=0}^{N-1} e^{z_k} - e^{z_i} e^{z_j}}{(\sum_{k=0}^{N-1} e^{z_k})^2}$$

$$= \frac{e^{z_i}}{\sum_{k=0}^{N-1} e^{z_k}} \cdot \frac{(\sum_{k=0}^{N-1} e^{z_k}) - e^{z_j}}{\sum_{k=0}^{N-1} e^{z_k}} = P_i(1 - P_j)$$

Similarly for the case of $i \neq j$ we get:

$$\frac{\partial P_i}{\partial z_j} = \frac{\partial \left( \frac{e^{z_i}}{\sum_{k=0}^{N-1} e^{z_k}} \right)}{\partial z_j} = \frac{0 - e^{z_i} e^{z_j}}{(\sum_{k=0}^{N-1} e^{z_k})^2}$$

$$= \frac{-e^{z_i}}{\sum_{k=0}^{N-1} e^{z_k}} \cdot \frac{e^{z_j}}{\sum_{k=0}^{N-1} e^{z_k}} = -P_i P_j$$

Combining the two last results we get:

$$\frac{\partial P_i}{\partial z_j} = \begin{cases} P_i(1 - P_j), & i = j \\ -P_i P_j, & i \neq j \end{cases}$$

Derivation of the Cross-Entropy Loss
The cross-entropy loss associated with the given input $x$ is defined as:

$$L = -\sum_{i=0}^{N-1} y_i \cdot log(P_i)$$

where $Y = (y_0, y_1, \ldots, y_{N-1})$ is the one-hot encoded ground truth vector ($y_i \in \{0,1\}$).

Treating 'log' as 'ln' for the sake of derivation, we can therefore express the gradient of the cross-entropy loss with respect to the i-th logit component as:

$$\frac{\partial L}{\partial z_i} = \frac{\partial(-\sum_{k=0}^{N-1} y_k \cdot log(P_k))}{\partial z_i}$$

$$= -\sum_{k=0}^{N-1} y_k \cdot \frac{\partial \, log(P_k)}{\partial z_i} = -\sum_{k=0}^{N-1} y_k \cdot \frac{\partial \, log(P_k)}{\partial P_k} \cdot \frac{\partial P_k}{\partial z_i}$$

$$= -\sum_{k=0}^{N-1} y_k \cdot \frac{1}{P_k} \cdot \frac{\partial P_k}{\partial z_i}$$

Combining the Softmax Function Cross-Entropy Derivatives
Given that the softmax derivative equation for the case when $i = j$ is different from all other cases, we reorder the loss derivative equation a little separating this case from the others:

$$\frac{\partial L}{\partial z_i} = -\sum_{k=0}^{N-1} y_k \cdot \frac{1}{P_k} \cdot \frac{\partial P_k}{\partial z_i}$$

$$= -y_i \cdot \frac{1}{P_i} \cdot \frac{\partial P_i}{\partial z_i} - \sum_{k \neq i} y_k \cdot \frac{1}{P_k} \cdot \frac{\partial P_k}{\partial z_i}$$

Now, we can use the softmax derivative we have calculated earlier and get:

$$= -y_i \cdot \frac{P_i(1 - P_i)}{P_i} - \sum_{k \neq i} \frac{y_k \cdot (-P_k P_i)}{P_k} = -y_i + y_i P_i + \sum_{k \neq i} y_k P_i$$

$$= P_i \left( y_i + \sum_{k \neq i} y_k \right) - y_i$$

However, since $Y$ is the one-hot encoded ground truth vector, we get that:

$$y_i + \sum_{k \neq i} y_k = \sum_{k=0}^{N-1} y_k = 1$$

and hence:

$$\frac{\partial L}{\partial z_i} = P_i \left( y_i + \sum_{k \neq i} Y_k \right) - y_i = P_i - y_i$$